# Workspace and Kinematic Analysis of the VERNE machine

Daniel KANAAN, Philippe WENGER and Damien CHABLAT

*Abstract*— **This paper describes the workspace and the inverse and direct kinematic analysis of the VERNE machine, a serial/parallel 5-axis machine tool designed by Fatronik for IRCCyN. This machine is composed of a three-degree-of-freedom (DOF) parallel module and a two-DOF serial tilting table. The parallel module consists of a moving platform that is connected to a fixed base by three non-identical legs. This feature involves (i) a simultaneous combination of rotation and translation for the moving platform, which is balanced by the tilting table and (ii) workspace whose shape and volume vary as a function of the tool length. This paper summarizes results obtained in the context of the European projects NEXT ("Next Generation of Productions Systems").**

## I. INTRODUCTION

PARALLEL kinematic machines (PKM's) are well known for their high structural rigidity, better payload-to-weight ratio, high dynamic performances and high accuracy [1-3]. Thus, they are prudently considered as attractive alternatives designs for demanding tasks such as high-speed machining [4].

A well-known feature of PKM is the existence of multiple solutions to the direct kinematic problem. That is, the moving platform can admit several positions and orientations (poses) in the workspace for one given set of input joint values [5]. Moreover, parallel manipulators exist with multiple inverse kinematic solutions. This means that the moving platform can admit several input joint values corresponding to one given pose of the end-effector [6]. For industrial PKMs, the resolution of the direct and inverse kinematic problems is usually made using iterative methods. With such methods, the calculation time can change as function of the configuration of the PKM. A major drawback of PKMs is a complex workspace. The workspace of a PKM has not a simple geometric shape, and its functional volume is reduced as compared to the space occupied by the machine [7].

D. Kanaan is with the Institut de Recherche en Communications et Cybernétique de Nantes (IRCCyN) (UMR CNRS 6597), France (corresponding author to provide phone: 33-240376958; fax: 33-240376930, e-mail: Daniel.Kanaan@irccyn-nantes.fr); P. Wenger is with the IRCCyN (UMR CNRS 6597), France (e-mail: Philippe.Wenger@irccyn-nantes.fr); D. Chablat is with the IRCCyN (UMR CNRS 6597), France (e-mail: Damien.Chablat@irccyn-nantes.fr).

Because many industrial tasks require less than six degrees of freedom, several lower-DOF PKMs have been developed. For some of theme, the reduction of the number of DOFs can result in coupled motions of the moving platform [8-10]. This is the case, for example, in the RPS manipulator [9] and in the parallel module of the VERNE machine. The kinematic modeling of these PKMs must be done case by case according to their structure.

Many researchers have contributed to the study of the kinematics of lower-DOF PKMs. Many of them have focused on the discussion of both analytical and numerical methods [11-12]. This paper investigates the workspace and the inverse and direct kinematics analysis of the VERNE machine (Fig. 1).

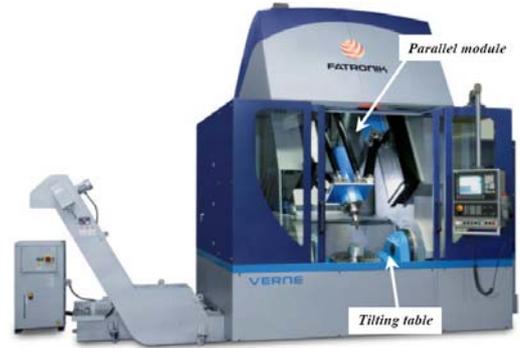

Fig. 1. Overall view of the VERNE machine

The following section describes the VERNE machine. In section III, we study the kinematics of the VERNE machine. In section IV, we present the workspace calculation for different tool length and for different orientation angle of the tool. Finally Section V concludes this paper.

## II. DESCRIPTION OF THE VERNE MACHINE

The VERNE machine consists of a parallel module and a tilting table as shown in Fig. 2. The vertices of the moving platform of the parallel module are connected to a fixed-base plate through three legs I, II and III. Each leg uses a pair of rods linking a prismatic joint to the moving platform through two pairs of spherical joints. Legs II and III are two identical parallelograms. Leg I differs from the other two legs in that $A_{11}A_{12} \neq B_{11}B_{12}$, where $A_{ij}$ (respectively

$B_{ij}$) is the center of spherical joint number j on the prismatic joint number i (respectively on the moving platform side), i = 1..3, j = 1..2. The movement of the moving platform is generated by three sliding actuators along three vertical guideways.

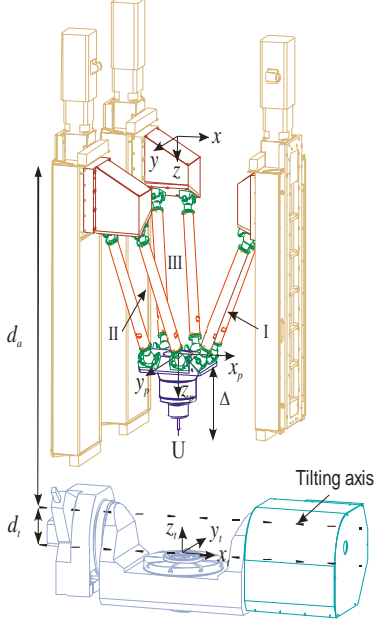

Fig. 2: Schematic representation of the VERNE machine

Due to the arrangement of the links and joints, legs II and III prevent the platform from rotating about y and z axes. Leg I prevents the platform from rotating about z-axis (Fig. 2) because $A_{11}A_{12} \neq B_{11}B_{12}$, however, a slight coupled rotation $\alpha$ about x-axis exists. The tilting table is used to rotate the workpiece about two orthogonal axes. The first one, the tilting axis, is horizontal and the second one, the rotary axis, is always perpendicular to the tilting table. This machine takes full advantage of these two additional axes to adjust the tool orientation with respect to the workpiece.

### III. KINEMATIC ANALYSIS OF THE VERNE MACHINE

#### A. Kinematic equations of the VERNE machine

In order to analyze the kinematics of the VERNE machine, three relative coordinates are assigned as shown in Fig. 2. A static Cartesian frame $R_b = (O, x, y, z)$ is fixed at the base of the machine tool, with the z-axis pointing downward along the vertical direction. Two mobile Cartesian frame, the first frame $R_{pl} = (P, x_P, y_P, z_P)$, is attached to the moving platform at point P and the second frame, $R_t(t, x_t, y_t, z_t)$ is attached to the tilting table at point t. Let us $^bT_{pl}$ define the transformation matrix that brings the fixed Cartesian frame $R_b$ on the frame $R_{pl}$ linked to the moving platform.

$$^bT_{pl} = Trans(x_p, y_p, z_p) Rot(x, \alpha) \quad (1)$$

We use this transformation matrix to express $B_{ij}$ as function of $x_p, y_p, z_p$ and $\alpha$ by using the relation $B_{ij} = {}^bT_{pl}{}^{pl}B_{ij}$ where $^{pl}B_{ij}$ represents the point $B_{ij}$ expressed in the frame $R_{pl}$.

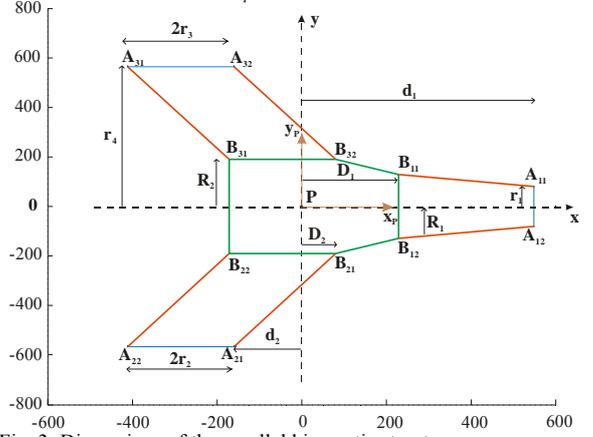

Fig. 3: Dimensions of the parallel kinematic structure

Using the parameters defined in Figs. 2 and 3, the constraint equations of the parallel manipulator are expressed as:

$$(x_{Bij} - x_{Aij})^2 + (y_{Bij} - y_{Aij})^2 +$$
$$(z_{Bij} - z_{Aij})^2 - L_i^2 = 0 \quad (i = 1..3, j = 1..2) \quad (2)$$

Leg I is represented by two different Eqs. (3-4). This is due to the fact that $A_{11}A_{12} \neq B_{11}B_{12}$ (figure 3).

$$(x_P + D_1 - d_1)^2 + (y_P + R_1 \cos(\alpha) - r_1)^2$$
$$+ (z_P + R_1 \sin(\alpha) - \rho_1)^2 - L_1^2 = 0 \quad (3)$$

$$(x_P + D_1 - d_1)^2 + (y_P - R_1 \cos(\alpha) + r_1)^2$$
$$+ (z_P - R_1 \sin(\alpha) - \rho_1)^2 - L_1^2 = 0 \quad (4)$$

Leg II is represented by a single Eq. (5).

$$(x_P + D_2 - d_2)^2 + (y_P - R_2 \cos(\alpha) + r_4)^2$$
$$+ (z_P - R_2 \sin(\alpha) - \rho_2)^2 - L_2^2 = 0 \quad (5)$$

The leg III, which is similar to leg II (figure 3), is also represented by a single Eq. (6).

$$(x_P + D_2 - d_2)^2 + (y_P + R_2 \cos(\alpha) - r_4)^2$$
$$+ (z_P + R_2 \sin(\alpha) - \rho_3)^2 - L_3^2 = 0 \quad (6)$$

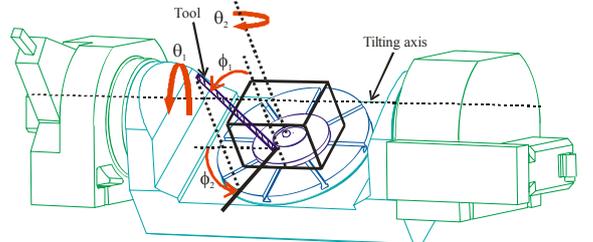

Fig. 4: Draw of the tilting table where the tool orientation is defined by ($\phi_1$, $\phi_2$) relative to $R_t$ and the orientation of the tilting table is defined by ($\theta_1$, $\theta_2$) relative to $R_b$

Let $^bT_t$ define the transformation matrix that brings the fixed Cartesian frame $R_b$ on the frame $R_t$ linked to the tilting table.

$$^bT_t = trans(z, d_a) rot(x, \theta_1) trans(z, d_t) rot(x, \pi) rot(z, \theta_2) \quad (7)$$

Let $^tT_{pl}$ define the transformation matrix that brings the frame $R_t$ linked to the tilting table on the frame $R_{pl}$ linked to the moving platform; where $X_u$, $Y_u$ and $Z_u$ are the coordinates of the tool centre point (TCP), U, in $R_t$.

$$^tT_{pl} = trans(X_u, Y_u, Z_u) rot(z, \phi_2) rot(x, \pi + \phi_1) trans(z, -\Delta)$$
(8)

We use transformation matrices from Eqs. (7) and (8) in order to express $B_{ij}$ as function of $X_u$, $Y_u$, $Z_u$, $\phi_1$, $\phi_2$, $\theta_1$ and $\theta_2$ by using the relation $B_{ij} = ^bT_{pl} \, ^{pl}B_{ij}$ where $^bT_{pl} = ^bT_t \, ^tT_{pl}$ and $^{pl}B_{ij}$ represents the point $B_{ij}$ expressed in the frame $R_{pl}$.

Using Eq. (2) and the parameters defined in Figs. 2, 3 and 4, we can express all constraint equations of the VERNE machine. However knowing that $A_{i1}B_{i1}$ and $A_{i2}B_{i2}$ are parallel for i=1..2, we can prove that:

$$\theta_2 = -\phi_2 \quad (9)$$

Substituting the above value of $\theta_2$ in all constraint equations resulting from Eq. (1), we obtain that leg I is represented by two different equations (10) and (11) while leg II (respectively leg III) is represented by only one equation (12) (respectively equation 13)).

Equations (10-13) are not reported here because of space limitation. They are available in [14].)

If we identify Eqs. (10), (11), (12) and (13) with Eqs. (3), (4), (5) and (6) respectively, we conclude that:

$$\alpha = \theta_1 + \phi_1 \quad (14)$$

The constraint equations of the VERNE machine will be used in order to obtain the inverse kinematic models of the full VERNE machine.

### B. Coupling between the position and the orientation of the platform

The parallel module of the VERNE machine possesses three actuators and three degrees of freedom. However, there is a coupling between the position and the orientation angle of the platform. The object of this subsection is to study the coupling constraint imposed by leg I.

By eliminating $\rho_1$ from Eqs. (3) and (4), we obtain a relation (15) between $x_P$, $y_P$ and $\alpha$ independently of $z_P$.

$$R_1^2 \sin^2(\alpha)(x_P + D_1 - d_1)^2 + (r_1^2 - 2R_1 r_1 \cos(\alpha) + R_1^2) y_P^2$$
$$-R_1^2 \sin^2(\alpha)\left(L_1^2 - (R_1^2 + r_1^2 - 2R_1 r_1 \cos(\alpha))\right) = 0 \quad (15)$$

We notice that for a given $\alpha$, Eq. (15) represents an ellipse (16). The size of this ellipse is determined by $a$ and $b$, where $a$ is the length of the semi major axis and $b$ is the length of the semi minor axis.

$$\frac{(x_P + D_1 - d_1)^2}{a^2} + \frac{y_P^2}{b^2} = 1 \quad (16)$$

where
$$\begin{cases} a = \sqrt{\left(L_1^2 - (R_1^2 + r_1^2 - 2R_1 r_1 \cos(\alpha))\right)} \\ b = \sqrt{\dfrac{R_1^2 \sin^2(\alpha)\left(L_1^2 - (R_1^2 + r_1^2 - 2R_1 r_1 \cos(\alpha))\right)}{(r_1^2 - 2R_1 r_1 \cos(\alpha) + R_1^2)}} \end{cases}$$

These ellipses define the locus of points reachable with the same orientation $\alpha$.

### C. The Inverse Kinematics

For the inverse kinematic problem of the parallel module, the position coordinates ($x_P$, $y_P$, $z_P$) are given but the coordinates $\rho_i$ ($i = 1..3$) of the actuated prismatic joints and the orientation angle $\alpha$ of the moving platform are unknown. The problem consists in solving the system ($S1$) of 4 equations (3-6) for only 4 unknowns ($\rho_i$ ($i = 1..3$) and $\alpha$). Thus, the position of the TCP ($X_u$, $Y_u$, $Z_u$) and the orientation of the tool ($\phi_1$ and $\phi_2$) are given relative to the frame $R_t$, but the joint coordinates, defined by the position $\rho_i$ ($i = 1..3$) of the actuated prismatic and the orientation ($\theta_1$ and $\theta_2$) of the tilting table in the base frame $R_b$ are unknown. Knowing that $\theta_2 = -\phi_2$ from Eq. (9), the problem consists in solving the system ($S2$) of 4 equations ((10), (11), (12) and (13)) for only 4 unknowns ($\rho_i$ ($i = 1..3$) and $\theta_1$).

In both cases, we follow the same reasoning to solve the inverse kinematics. Due to space limitation, we present only the inverse kinematic model of the full VERNE machine. The inverse kinematic model of the parallel module can be obtained easily by considering system ($S1$) instead of system ($S2$) (see [13]).

First, we eliminate $\rho_1$ from Eqs. (10) and (11) in order to obtain a relation (17) between the TCP position and orientation ($X_u$, $Y_u$, $Z_u$, $\phi_1$, $\phi_2$) and the tilting angle $\theta_1$. Then, we find all possible orientation angles $\theta_1$ for prescribed values of the position and the orientation of the tool. These orientations are determined by solving a six-degree-characteristic polynomial in $\tan(\theta_1/2)$ derived from Eq. (17). This polynomial can have up to four real solutions. This conclusion is verified by the fact that $\theta_1 = \phi_1 - \alpha$ from Eq. 30 where $\alpha$ can have only four real solutions (which can be proved by drawing together all ellipses of iso-values of $\alpha$, see [13]). After finding all the possible orientations, we use the system of equations ($S2$) to calculate the joint coordinates $\rho_i$ for each orientation angle $\theta_1$.

For $\rho_1$, we must verify that the values of $\rho_1$ obtained from Eqs. (10) and (11) are the same. As a result, we eliminate one of the two solutions.

Observing the above remark and knowing from [16] that Eqs (10-11), (12), (13) are defined as two-degree-polynomials in $\rho_i, i=1..3$ respectively, we conclude that there are four solutions for leg I and two solutions for leg II and III. Thus there are sixteen inverse kinematic solutions for the VERNE machine.

From the sixteen theoretical inverse kinematics solutions shown in Fig. 5, only one is used by the VERNE machine: the one referred to as (p) in Fig. 5, which is characterized by the fact that each leg must have its slider attachment points above the moving platform attachment points.

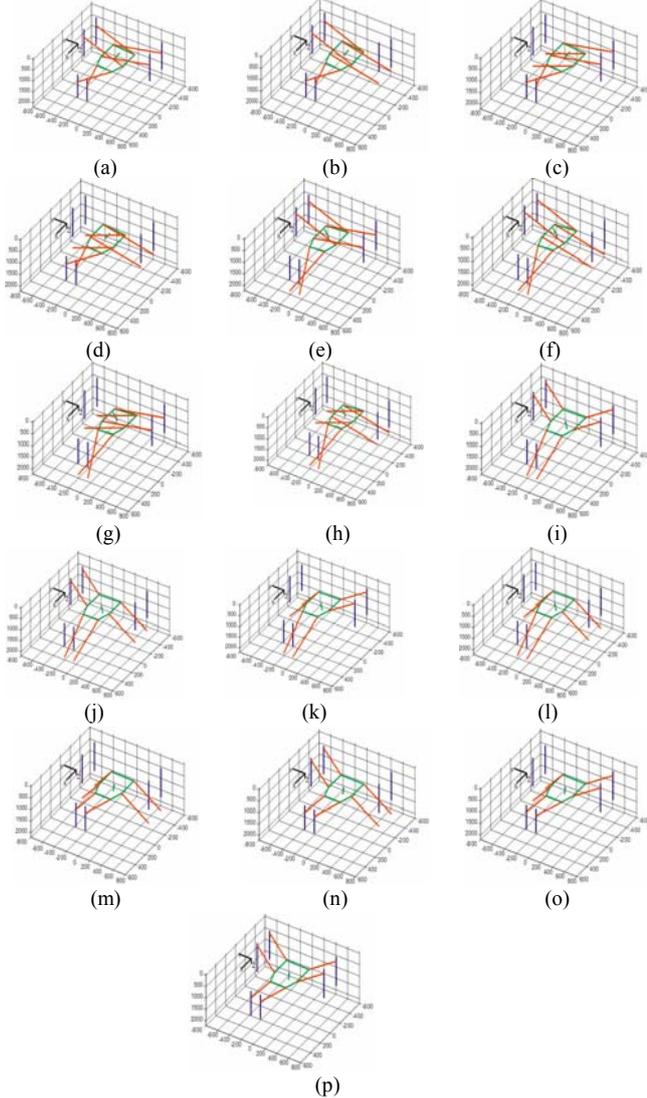

Fig 5: The sixteen solutions to the inverse kinematics problem when $x_P = -240$ mm, $y_P = -86$ mm and $z_P = 1000$ mm

For the remaining 15 solutions one of the sliders leaves its joint limits or the two rods of leg I cross. Most of these solutions are characterized by the fact that at least one of the legs has its slider attachment points lower than the moving platform attachment points. To prevent rod crossing, we also add a condition on the orientation of the moving platform. This condition is $R_1 \cos(\theta_1 + \phi_1) > r_1$. Finally, we check the joint limits of the sliders and the serial singularities [10].

Applying the above conditions will always yield a unique solution for practical applications. The proposed method for calculating the various solutions of the inverse kinematic problem has been implemented in C++. We are working with AMTRI, a UK industry, on implementing this code in a simulation package of PKMs, Visual Components.

### D. The Forward Kinematics

For the forward kinematics of our spatial parallel manipulator, the values of the joint coordinates $\rho_i$ ($i=1..3$) are known and the goal is to find the coordinates $x_P$, $y_P$ and $z_P$ of the centre of the moving platform P.

To solve the forward kinematics, we successively eliminate variables $x_P$, $y_P$ and $z_P$ from the system ($S1$) of four equations (3-6) to lead to an equation function of the joint coordinates $\rho_i$ ($i=1..3$) and function of the orientation angle $\alpha$ of the platform. To do so, we first compute $y_P$ as function of $z_P$ by subtracting equation (3) from equation (4) and we replace this variable in system ($S1$) to obtain a new system ($S3$) of three equations (18), (19) and (20) derived from equations (3), (5) and (6) respectively. We then compute $z_P$ as function of $\rho_i$ ($i=1..3$) and $\alpha$ by subtracting equation (19) from equation (20). We replace this variable in system ($S3$) to obtain a new system ($S4$) of two equations (21) and (22) derived from equations (18) and (19) respectively. Finally, we compute $x_P$ as function of $\rho_i$ ($i=1..3$) and $\alpha$ by subtracting equation (21) from equation (22) and we replace this variable in the system ($S4$) in order to eliminate $x_P$. Equations of system ($Si$) (i=3..4) are not reported here because of space limitation. They are available in [14].

For each step, we determine solutions existence conditions by studying the denominators that appear in the expressions of $x_P$, $y_P$ and $z_P$. These conditions are:

$$R_1 \cos(\alpha) - r_1 \neq 0 \quad (23)$$

$$(\rho_2 - \rho_3)(R_1 \cos(\alpha) - r_1) + 2\sin(\alpha)(r_4 R_1 - r_1 R_2) \neq 0 \quad (24)$$

Equation (23) implies that $A_1 B_1$ is perpendicular to the slider plane of leg I. In this case equation (16) represents a circle because $a=b$.

When $\rho_2 = \rho_3$ in equation (24), we have $\alpha = \{0, \pi\}$. This means that $y_P = 0$ (obtained from Equations. (5)–(6)).

To finish the resolution of the system, we perform the tangent-half-angle substitution $s = \tan(\alpha/2)$. As a consequence, the forward kinematics of our parallel manipulator results in a eight-degree-characteristic polynomial in $s$, whose coefficients are relatively large expressions in $\rho_1$, $\rho_2$ and $\rho_3$.

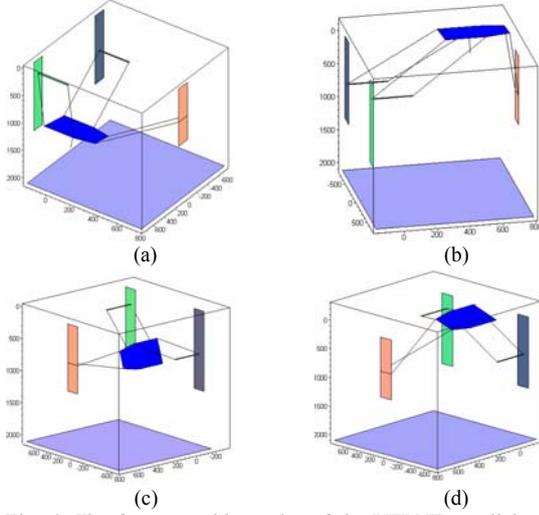

Fig. 6: The four assembly-modes of the VERNE parallel module for $\rho_1 = 674$ mm, $\rho_2 = 685$ mm and $\rho_3 = 250$ mm. only (a) is reachable by the actual machine

For the forward kinematics of the VERNE machine, the values of the joint coordinates, defined by the position $\rho_i$ ($i=1..3$) of the actuated prismatic and the orientation ($\theta_1$ and $\theta_2$) of the tilting table in the base frame $R_b$ are known and the goal is to find the position of the TCP ($X_u$, $Y_u$, $Z_u$) and the orientation of the tool ($\phi_1$ and $\phi_2$) in the frame $R_t$.

Knowing that $\phi_2 = -\theta_2$ and $\phi_1 = \alpha - \theta_1$ from Eqs (9) and (14), we solve this problem by first solving the forward kinematics of the parallel module of the VERNE machine in order to find the coordinates $x_P$, $y_P$ and $z_P$ of the centre of the moving platform P and the orientation $\alpha$ of the moving platform in term of the joint coordinates $\rho_i$ ($i=1..3$). We then use transformation matrices from Eqs. (1) and (7) in order to express the tool position and orientation ($X_u$, $Y_u$, $Z_u$, $\phi_1$ and $\phi_2$) as function of $(x_P, y_P, z_P, \theta_1, \theta_2)$.

$$^tU = {^tT_b}{^bU} = {^tT_b}{^bT_{pl}}{^{pl}U} = {^bT_t^{-1}}{^bT_{pl}}{^{pl}U} \quad (25)$$

where $^{pl}U = \begin{bmatrix} 0 & 0 & \Delta & 1 \end{bmatrix}^T$ and $^tU = \begin{bmatrix} X_u & Y_u & Z_u & 1 \end{bmatrix}^T$ represent the TCP, $U$, expressed in frames $R_{pl}$ (linked to the moving platform) and the base frame $R_b$ respectively. Finally we obtain:

$$\begin{cases} \phi_1 = \alpha - \theta_1 \text{ and } \phi_2 = -\theta_2 \\ X_u = \cos(\theta_2)x_p + V1\sin(\theta_2) \end{cases} \begin{cases} Y_u = -\sin(\theta_2)x_p + V1\cos(\theta_2) \\ Z_u = \sin(\theta_1)y_p + V2 \end{cases}$$
(26)

where $V1 = \Delta\sin(\alpha - \theta_1) - \cos(\theta_1)y_p - \sin(\theta_1)(z_p - d_a)$ and $V2 = d_t - \cos(\theta_1)z_p + d_a\cos(\theta_1) - \Delta\cos(\alpha - \theta_1)$

For the VERNE machine, only 4 assembly-modes have been found (figure 6). It was possible to find up to 6 assembly-modes but only for input joint values out of the reachable joint space of the machine. Only one assembly-mode is actually reachable by the machine (solution (a) shown in Fig. 6) because the other ones lead to either rod crossing, collisions, or joint limit violation. The right assembly mode can be recognized, like for the right working mode, by the fact that each leg must have its slider attachment points above the moving platform attachment points

The proposed method for calculating the various solutions of the forward kinematic problem has been implemented in Maple.

IV. WORKSPACE CALCULATION OF THE VERNE MACHINE

A. *Workspace calculation of the parallel module [10]*

The parallel module of the VERNE machine possesses 3 degrees of freedom. A complete representation of the workspace is a volume. Consequently this workspace can be defined by the positions in space reachable by the point P, centre of the mobile Cartesian frame $R_{pl}$. However the parallel module undergoes a complex motion, thus to overcome the complexity caused by the presence of the coupled rotation, we propose the following method.

Step 1. We virtually cut the leg I, which is constituted of rods 11 and 12 (rod ij denotes $A_{ij}B_{ij}$) by supposing that $\rho_{11}$ is independent of $\rho_{12}$, where $\rho_{11}$ and $\rho_{12}$ are respectively the joint coordinates of the prismatic joints linking rods 11 and 12 to their guideways. The parallel module of the VERNE machine possesses now 4 degrees of freedom instead of 3. These degrees of freedom are defined by coordinates $x_P$, $y_P$ and $z_P$ of the point P and by the orientation $\alpha$ of the moving platform. Thus we consider that the orientation $\alpha$ of the platform is given, and we geometrically model constraints limiting the workspace of the new parallel architecture. These constraints are: (i) Interference between links, (ii) Leg Length limits, (iii) Serial Singularity, (iv) Mechanical limits on passive joints (we take into consideration the type of joints as well as the location of these joints in the machine) and (v) Actuator stroke. The intersection between these models is a volume.

Step 2. We consider the interdependence between rods 11 and 12 of leg I characterized by the fact that $\rho_{11} = \rho_{12} = \rho_1$. This will allows us to determine the geometric shape for which the coupling between the position and the orientation of the platform exists. Thus for a given orientation $\alpha$, the point P describes a surface defined by a hollow cylinder whose base is an ellipse as shown in subsection III.B.

The intersection between geometric models defined at step 1 and the surface defined at step 2 represents the constant orientation workspace of the new 4 degree-of-freedom parallel module. We calculate a horizontal cut of this workspace when the point P, center of the mobile Cartesian frame, moves in a known horizontal plane. Then we proceed by discretization to determine the complete workspace of the parallel module of the VERNE machine (see Figure 7). To calculate the workspace for a given tool length, we only need to express the TCP, U, in the base frame as function of $(x_P, y_P, z_P, \alpha)$ by using the transformation matrix from Eq. (1), $U = {}^b T_{pl}{}^{pl}U$:

$$x_U = x_P,\ y_U = y_P - \Delta \sin(\alpha) \text{ and } z_U = z_P + \Delta \cos(\alpha) \quad (27)$$

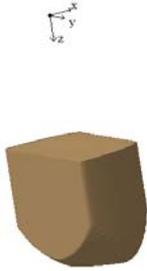
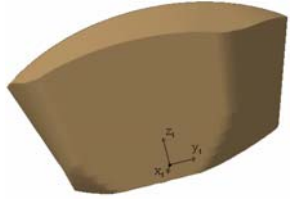

Fig. 7: Workspace defined as the set of points P in the fixed Cartesian frame $R_b$

Fig 8: The manufacturing 3D workspace for given tool length of 50 mm

### B. The manufacturing 3D workspace [14]

In this subsection, we calculate the manufacturing 3D workspace of the VERNE (the tool axis remains perpendicular to the part base) by considering that the tilting table rotates about its axis and follows the orientation $\alpha$ of the moving platform, which means that $\theta_1 = \alpha$ and $\phi_1 = 0$, we suppose also that $\theta_2 = \phi_2 = 0$.

To calculate this workspace we use Eq. (26) after replacing $\theta_1$, $\theta_2$, $\phi_1$, and $\phi_2$ by their values. This will give us the position and the orientation of the TCP in the frame $R_t$ ($X_u$, $Y_u$, $Z_u$, $\phi_1$, $\phi_2$) expressed as function of ($x_p$, $y_p$, $z_p$, $\alpha$). A graphical representation of the workspace of PKMs with more than three degrees of freedom is only possible if we fix parameters representing the exceeded degrees of freedom. Thus to calculate the workspace for various orientations of the tool, we first fix $\phi_1$ and $\phi_2$, then we calculate the position of the TCP ($X_u$, $Y_u$, $Z_u$) in the frame $R_t$ by using Eq. (26).

The proposed method for calculating the workspace for various tool lengths and for different tool orientation angles has been implemented in Maple and displayed in the CAD software CATIA. The obtained results show a workspace larger than the one currently used by the VERNE machine (for some given tool orientation angles). So this will able us to improve the productivity of the VERNE machine and to reach the limit of its capacity without risk of collision or damage of the VERNE machine.

## V. CONCLUSIONS

This paper was devoted to the kinematic and workspace analysis of a 5-DOF hybrid machine tool, the VERNE machine. This machine possesses a complex motion caused by the unsymmetrical architecture of the parallel module where one of the legs is different from the other two legs. It was shown that the inverse kinematics has sixteen solutions and the forward kinematics may have six real solutions. The workspace was calculated by using a combination of discretization and geometrical method. This work is of interest as it may improve the efficiency of the machine because the controller of the actual VERNE machine resorts to an iterative Newton-Raphson resolution of the kinematic models.

### ACKNOWLEDGMENT

This work has been partially funded by the European projects NEXT, acronyms for "Next Generation of Productions Systems", Project no° IP 011815. The authors would like to thank the Fatronik society, which permitted us to use the CAD drawing of the Machine VERNE what allowed us to present well the machine.